# Causality, Causal Discovery, and Causal Inference in Structural Engineering

M.Z. Naser, PhD, PE
School of Civil & Environmental Engineering and Earth Sciences (SCEEES), Clemson University, USA
Artificial Intelligence Research Institute for Science and Engineering (AIRISE), Clemson University, USA
E-mail: mznaser@clemson.edu, Website: www.mznaser.com

**Abstract**
Much of our experiments are designed to uncover the cause(s) and effect(s) behind a data generating mechanism (i.e., phenomenon) we happen to be interested in. Uncovering such relationships allows us to identify the *true* working of a phenomenon and, most importantly, articulate a model that may enable us to further explore the phenomenon on hand and/or allow us to predict it accurately. Fundamentally, such models are likely to be derived via a *causal* approach (as opposed to an observational or empirical mean). In this approach, causal *discovery* is required to create a causal model, which can then be applied to *infer* the influence of interventions, and answer any hypothetical questions (i.e., in the form of What ifs? Etc.) that we might have. This paper builds a case for causal discovery and causal inference and contrasts that against traditional machine learning approaches – all from a civil and structural engineering perspective. More specifically, this paper outlines the key principles of causality and the most commonly used algorithms and packages for causal discovery and causal inference. Finally, this paper also presents a series of examples and case studies of how causal concepts can be adopted for our domain.

<u>*Keywords*</u>: Causal discovery; Causal inference; Civil engineering; Machine learning.

**Introduction**
Seeking causal knowledge is a foundational pursuit with branching philosophical, epistemological, and ontological ties[1]. Causal knowledge accurately describes how a phenomenon, *Y*, comes to be by answering key questions such as, what causes *Y*? How, when, and why does *Y* occur? What would happen if we were to intervene and influence *Y* in a certain way? Or perhaps, one of *Y*'s governing factors? Etc. [1]. Arriving at complete and precious answers to the above questions is a hefty and ambitious endeavor, and thus, causal knowledge can be difficult to pinpoint, measure, or obtain [2].

In our domain, we often design experiments in which a set of parameters is identified to be responsible for a given phenomenon. For example, say we are trying to understand the concept of flexural capacity in beams. First, we would fabricate a Beam A with certain geometrical and material features (i.e., W16×36, Grade A992). Then, we add boundary conditions (i.e., simply supported) to load this beam in a manner that enables us to capture its flexural response. We load the beam[2] and report that this beam fails once the level of the applied moment reaches 361.6 kN.m.

At this point, a link is then obtained by associating the reported moment at failure to the geometrical and material features as well as the loading configuration of Beam A. This link draws from the following observation: applying a bending moment of 361.6 kN.m to Beam A (of the above features) has <u>caused</u> the beam to fail. A series of questions may arise: 1) what has caused the failure of Beam A? And 2) was failure triggered due to the geometric configurations of W16×36? Or, perhaps due to the material properties of Grade A992? What about the effect of boundary conditions? Or Loading configuration?

---

[1] For a more cohesive look at causality from the intersection of philosophy, epistemology, and ontology, please refer to [44,45].
[2] Say with one point load at mid-span.

The above are examples of causal questions. Another set of questions that may also arise includes: other things constant [*Ceteris Paribus*] 3) would Beam A have failed if it was not for the presence of the bending moment? 4) would this beam fail at 361.6 kN.m if it had been a W18×40? Or had it been made from Grade A36? These are counterfactual questions that also belong to the causal family. Answering the above two sets of questions requires a causal investigation.

Fortunately, our domain knowledge can aid, if not substitute, the need for a thorough causal investigation. For example, we know that in the case of W-shaped steel beams under bending, the geometric features are lumped into the plastic modulus, $Z$, and that the material features are represented in the yield strength of steel, $f_y$. Both are also tied into an equation (see Eq. 1) of a multiplication form to estimate the resistance (ultimate moment capacity) of a given W-shaped steel beam.

*Resistance* = $Z \times f_y$ (1)

Equation 1 presents the simple fact that the flexural resistance of a W-shaped steel beam is a function of $Z$ and $f_y$. This equation also conveys that both $Z$ and $f_y$ contribute to the moment capacity in a complimentary (equal) manner. In other words, this equation allows us to answer all the above four questions intuitively.

For instance:
1) What has caused the failure of Beam A? *Ans. The presence of a bending moment that exceeds the level of moment capacity* – and hence the equality.

2) Was its failure triggered due to the geometric configurations of W16×36? Or, perhaps due to the material properties of Grade A992? *Ans. Building on Eq. 1, both Z and $f_y$ have contributed equally.*

3) *Ceteris Paribus*, could Beam A has failed if it was not for the presence of the bending moment? *Ans. No. Since if it was not for such loading, the capacity of the beam would not have been exceeded.*

4) *Ceteris Paribus*, Would this beam fail at 361.6 kN.m if it had been a W18×40? Or had it been made from Grade A36? *Ans. No. Since W18×40 has a larger plastic modulus than W16×36, and Grade 36 has lower yield strength than Grade A992.*

The identification of failure as outlined above may be perceived as a trivial problem[3]. In reality, it is a simple problem, and this simplicity stems from the narrow space of parameters involved in the phenomenon of flexural failure in W-shaped steel beams. However, the subject of causality exponentially grows with the addition of higher dimensions (say, flexural failure of tapered beams, beams subjected to high levels of shear-&-moment, presence of geometric imperfections and instability, etc.). The same is equally valid, if not more critical, for phenomena we do not have a working model (or, more appropriately, an/set of equation(s)) or those we lack domain knowledge of. Overcoming these critical limitations is one of the key motivations behind this work.

---

[3] A new question can also be formulated: 5) how could we have prevented failure in Beam A? *Ans. Failure can be prevented by switching Beam A with a larger section, and/or using a higher Grade of structural steel.* On a more fundamental and philosophical notion, *failure could have been prevented by reducing the level of bending moment, or possibly eliminating the existence of such bending moment.* The two former solutions stem from our engineering knowledge, while the latter stems from a causal perspective to our engineering understanding of the flexural failure problem.

A complimentary motivation stems from the fact that there is a very limited body of works that explored causality from a structural engineering perspective. Figure 1 presents the number of available studies that contain the terms [causality], [causal discovery], and [causal inference] in our domain over the last two decades. This figure was obtained through a scientometrics analysis from the open-source scholarly database, *Dimensions* [3,4]. As one can see, this analysis returns 27 papers, therefore averaging 1.35 papers per year. On a more positive note, there exist some works pertaining to causality in civil engineering sub-domains with a substantial human/social component, such as transportation and construction management [5–7]. Collectively, the number of studies on the front of causality from a civil engineering perspective is minute.

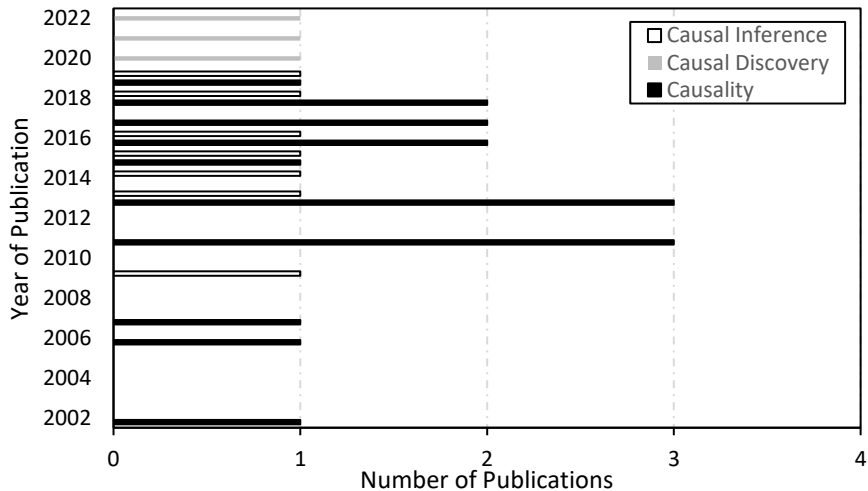

Fig. 1 Results of a scientometrics analysis on [causality], [causal discovery], and [causal inference] in structural engineering spanning 2002-2022.

In our pursuit of advancing knowledge, we seek to identify, or possibly retrieve, the underlying data generating process (DGP) responsible for the observations that result from the phenomenon we hope to understand [8]. As such, we devise experiments. Such experiments are designed to test and explore hypotheses. A hypothesis targets a specific direction to uncover the DGP behind a given phenomenon (i.e., the cause(s) leading to the so-called effect). In other words, our experiments are fueled by hypotheses that examine how a set of events/processes/states may produce other events/processes/states. More generally, cause(s) → effect, for, without the cause(s), the event would not have taken place [9].

Thus, this particular paper hopes to showcase the importance of re-visiting the principles of causality from a civil and structural engineering point of view. As such, this paper reviews key concepts pertaining to causality, causal discovery, and causal inference. Then, this paper outlines how causal machine learning and recent advancements in automated algorithms. To reinforce the covered concepts, this paper presents a series of examples and two case studies to establish causality in structural engineering problems. Finally, this paper ends with a discussion on how traditional machine learning can benefit from incorporating causal principles.

**Causality**

*Correlation does not imply causation*

Causality is the science of cause and effect; specifically, causality seeks to identify how effects (or events) come to be (or are caused) by their causes (or triggers). More often, causality is tied to correlation. However, correlation alone does *not* imply causation[4].

A prime example of spurious correlation can be seen in Fig. 2a and Fig. 2b, which depict a historical analysis of the number of failed buildings and bridges in the US as reported in [10,11] and compare that to the number of received PhDs in civil engineering (as per the National Science Foundation [12]). Figure 2a shows that there is a strong positive correlation (+84.9%) between the number of failed buildings in the US and the number of awarded PhDs in civil engineering between 1991-1996. Surprisingly, this plot may mistakenly imply that uptake in PhD graduates in our domain leads to (i.e., causes, or is somehow tied to) a rise in structural failures. Fortunately, this is not correct.

This correlation drops to a weak +20.6% when the number of awarded PhDs is contrasted against the number of failed bridges within the same timeframe (see Fig. 2b). This plot supports the notion that more PhD graduates in civil engineering do not strongly correlate with a rise in bridge failures. This interpretation is also not correct. Now, plotting the number of failed buildings and bridges in the US against the number of received PhDs in civil engineering between 1989-2000 yields a weak correlation of +35.5% and +3.87%, respectively (see Fig. 2c).

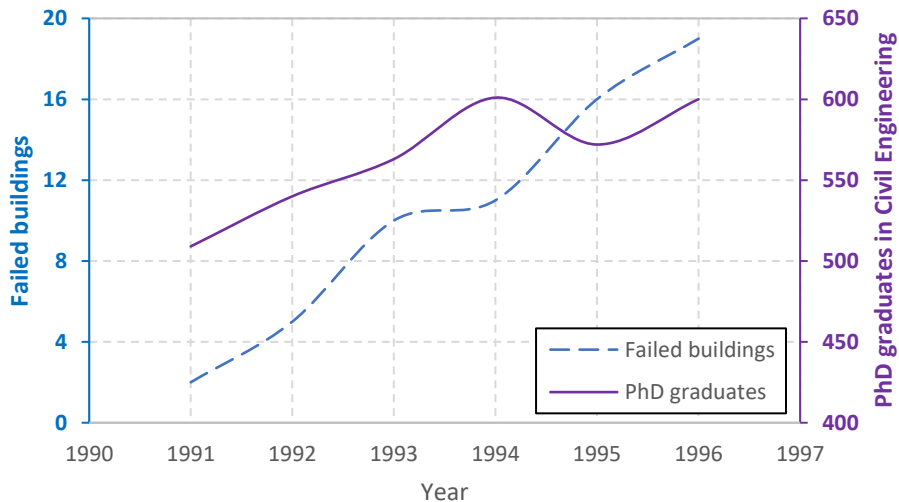

(a) Correlation between failed buildings in the US and the number of received PhDs in civil engineering within 1991-1996 (R = +84.9%)

---

[4] However, causation may imply some form of correlation [1].

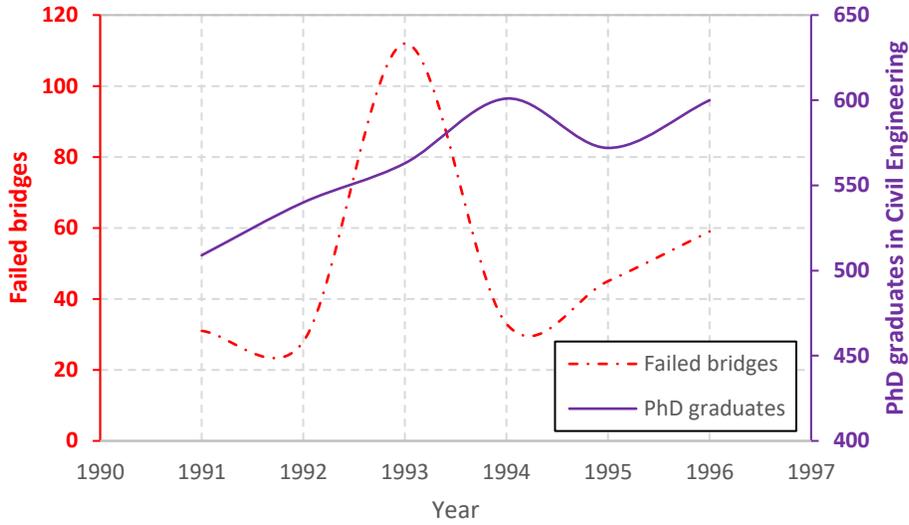

(b) Correlation between failed bridges in the US and the number of received PhDs in civil engineering within 1991-1996 (R = +20.6%)

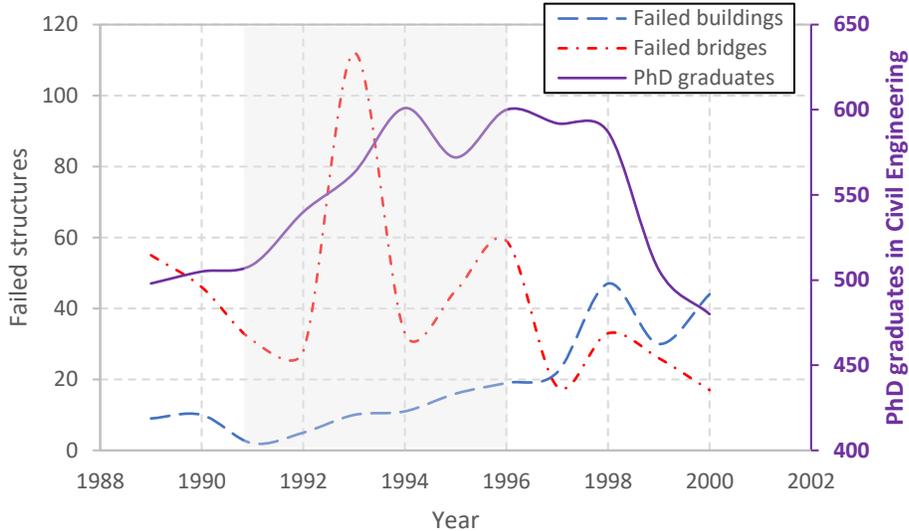

(c) Correlation between failed structures in the US and the number of received PhDs in civil engineering between 1989-2000 ($R_{buildings}$ = +35.6%, $R_{brigdes}$ = +3.87%)

Fig. 2 Examples of correlations

The above is but one example of why correlations alone do not entail causation simply because statistical relations do not exclusively confine causal relations. In fact, the presented correlations herein, as well as those that could be derived from Fig. 2[5], represent *marginal* associations (i.e., two variables are independent while ignoring a third) as opposed to *conditional* associations (e.g., two variables are independent <u>given</u> a third) and hence are unlikely to have a definite causal effect.

---

[5] For completion: substituting the outlier in 1993 in Fig. 2b by 40 failures, yields a strong 89.2% positive correlation between the number of awarded PhDs is contrasted against the number of failed bridges! Also, the correlation between these two items in the period of 1998-2000 is +93.6%.

While this particular example demonstrates shocking claims (yet, supported by authenticated data), let us look at a more practical example.

Figure 3 reinforces the difference between marginal associations and conditional associations in a more related demonstration. This figure examines the relationship between the applied loads on fire-exposed reinforced concrete (RC) columns against the compressive strength of concrete in each corresponding column. Figure 3a shows that the marginal dependence[6] between these two variables is positively linear, where columns subjected to relatively larger loads are also made from higher strength concrete (i.e., an increase in loading ($X$) co-occurs with an increase in compressive strength ($Y$)). However, when these two variables are conditioned on a third variable (occurrence of spalling, $Z$, see Fig. 3b), then one can clearly see that $X$ is independent ($\perp\!\!\!\perp$) of $Y$ given the occurrence of spalling ($Z$), as spalling seems to occur regardless of any marginal association between the applied loading and strength of concrete.

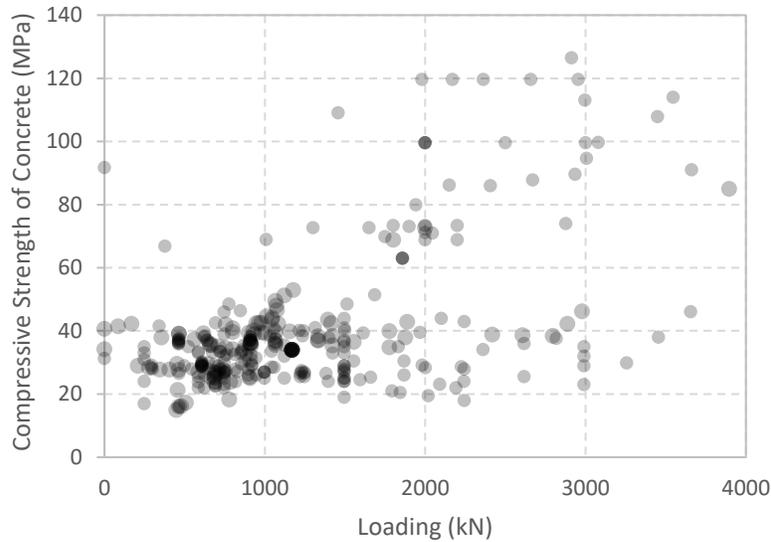

(a) Marginal dependence

---

[6] Simply, marginal independence implies that knowledge of $Y$'s value does not affect our belief in the value of $X$. Also, conditional independence implies that knowledge of $Y$'s value does not affect our belief in the value of $X$, given a value of $Z$.

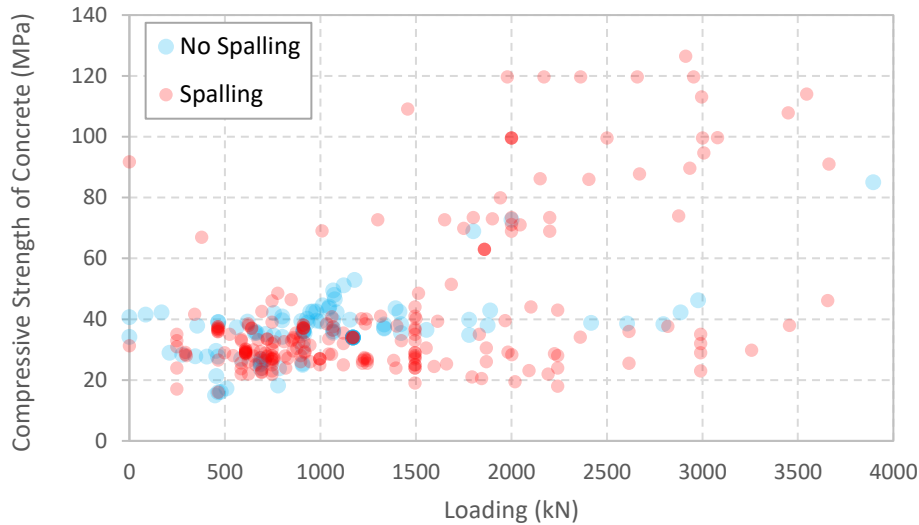

(b) Conditional <u>in</u>dependence

Fig. 3 Marginal vs. conditional association between the compressive strength of concrete and loading in fire-exposed RC columns

*The causal ladder*
This section outlines some of the key definitions of causality accepted in the open literature and then articulates key assumptions and algorithms often adopted to establish causation[7]. A good start to causality is to discuss how causality differs from traditional statistical and empirical methods, often favored by our domain. Our discussion will revolve around the causality rung system (or, more formally, the ladder of causation) pioneered by Pearl[8] [13]. In this system, Pearl identifies three levels (or, more formally, rungs) to causality.

The first rung builds upon the observations we *see* and hence stems from pure *statistical* associations. For example, an associational exercise would simply be the expectation that an uncoated metallic load bearing element (say, a plate girder) experience some level of corrosion after a few years of service. In this event, our observations of how metallic structures are likely to corrode come in handy to lay some form of association. Thus, such an association can answer the following simple question, *what is the expected level of corrosion an uncoated steel plate girder could undergo after 10 years of service?*

Mathematically, this associational question resembles the conditional probability of the occurrence of corrosion ($y$), given that we observed an uncoated steel plate girder ($x$), or P($y \mid x$). An experienced engineer may be able to provide us with a good and possibly accurate estimation of the expected magnitude of such corrosion if we are to provide this engineer with a list of information pertaining to the grade of steel, surrounding environmental conditions, etc. It is in this rung that *statistical* inference and empirical analysis reside. It is also in this rung we can build predictive models to predict such a phenomenon solely from data[9].

---
[7] In all cases, we confine this discussion to aspects relating to structural engineering. For completion, a philosophical and historical discussion and a more in-depth review on causality can be found elsewhere [15,22] and [46], respectively. In addition, this paper tries to limit the amount of mathematical background behind causality to maintain a smooth flow; noting that dedicated works on the mathematical conditions and background required for establishing causality are readily available in notable textbooks [22,34] and very recent review papers [19,31,47]. A discussion on such items (including do-calculus, conditional/interventional distributions, etc.), while worthy of our time and landscape of this study, is deemed too technical for an introductory work on causality in this domain.
[8] For a discussion on causality from the lens of the potential outcomes (PO), please refer to the following [34,48]. For brevity, we will discuss the PO approach in a later section.

The second rung of causality extends further than associations (the things we see) and includes *interventions* (the things we can *do*). Interventions are tasks that we *do* or *carry out* to explore questions such as what will happen to the system *if* we change one or more of its variables?

Let us continue our plate girder example from an interventional perspective. To do so, we will devise an experiment where we intervene on this girder to see how a particular variable we identify with having a causal relation (say, the addition of anti-corrosion coating) may or may not affect the corrosion of this girder. Mathematically, this interventional question resembles the use of do-calculus $P(y|do(x))$, which can be translated as the probability of event *y* (occurrence of corrosion) given that we intervene and set the value of *x* (addition of coating) and subsequently observe the same event (10 years of installation)[10].

As one can see, an accurate and quantitative answer to this question cannot be answered from data alone – unless we have previous results from an earlier experiment or, more conveniently, an accompanying new experiment in which we test (i.e., intervene) on a new girder. The same can also be said for machine learning (ML) enthusiasts. At this point in time, most supervised ML approaches are applied in a similar manner to that shown in Fig. 4a where a number of independent variables ($X_1$, $X_2$, $X_3$, …) is used to predict an outcome (*Y*). In contrast, *Y* occurs via DGP in a possible manner similar to Fig. 4b. Thus, we need to identify the DGP to accurately predict *Y*. This crucial difference distinguishes between a predictive statistical model and an interventional model.

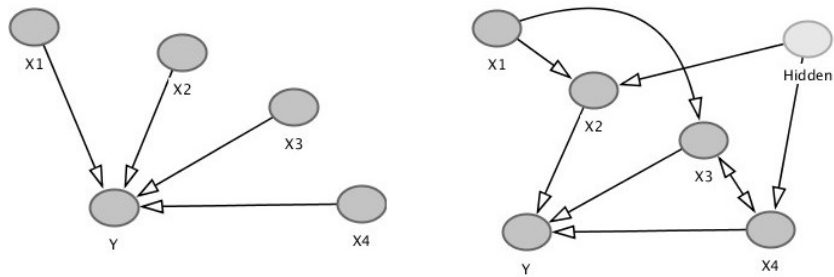

Fig. 4 Difference between a traditional ML model (left) and a causal model (right)

The third rung builds on the former two rungs to build counterfactual reasoning of causality. This rung is dedicated to answering questions of counterfactuals (the things we can *imagine*). For example, what would have happened to the expected level of corrosion of our plate girder after 15 years of exposure had this <u>particular</u> girder been fabricated from aluminum? This question could be translated to P($y_x$|$x'$, $y'$). Namely, the probability that event $Y = y$ would be observed had $X$ been $x$, given that we observed $X$ to be $x'$ and $Y$ to be $y'$. This question is not empirically testable since we cannot test the *exact same girder* twice, especially if it is fabricated from two different materials and examined in <u>two</u> different <u>exposures</u>. In other words, counterfactuals aim to answer individual-unit questions (while interventions answer population-level questions [14]).

---

[9] Just as many machine learning model do!
[10] There is a subtle difference between $P(y | x)$ and $P(y|do(x))$ wherein the former describes which values *Y* would likely take on when *X* happened to be *x* (i.e., observational distribution), while the latter describes the values of *y* when *X* is set to *x* (i.e., interventional distribution). These are not equal since the coated girder is likely to experience minimal corrosion.

A logical solution would be to run two experiments on identical girders (where one is fabricated from steel and the other from aluminum). Such an experiment is likely to give us a realistic answer to our counterfactual question. Fundamentally and philosophically, this may solve our query despite the fact that we did not test the exact same girder under different conditions. Yet, one can think of many examples were running an experiment may not be possible or ethical (e.g., would have happened to the occupants of a collapsed building, $X$, had the earthquake been twice as harsh? What would have happened to a community, $V$, had its buildings been designed for wildfire effects? etc.). This former example, among many others, emphasizes the need for a causal approach.

As one can see, while our domain incorporates all three rungs, we tend to heavily prioritize the first rung with a few ideas from the second (mechanically). Little has been paid to identifying the causal structure of DGP in our domain. While this author opts not to speculate on the above, two notions come to mind. It is possible that: 1) identifying DGPs is not only a complex endeavor but can be labeled as ambitious too, and 2) may not be needed since methods spanning from the first two rungs are suitable to provide us with <u>solutions</u> for our problems. This work hopes to present a case for its readers and our engineers on the value of seeking causality.

*Definitions and big ideas*
An early definition of causation is one established by Hume [15]. Hume defines causality from two perspectives, as a philosophical relation and a natural relation. In the former, a cause is "*an object precedent and contiguous to another, and where all objects resembling the former are placed in like relations of precedency and contiguity to those objects that resemble the latter*". On the other hand, the latter states that a cause is *"… an object precedent and contiguous to another, and so united with it that the idea of the one determines the mind to form the idea of the other, and the impression of the one to form a more lively idea of the other*". A simpler definition that is also attributed to Hume is also provided herein, "*We may define a cause to be an object, followed by another, [. . . ] where, if the first object <u>had not been</u>, the second <u>had never</u> existed*". The second part of this definition also describes a key companion concept to *causality*, which is *counterfactual*.

In terms of counterfactuals, Lewis [16] defines causality by, "*We think of a cause as something that makes a difference, and the difference it makes must be a difference from what would have happened without it. Had it been absent, its effects – some of them, at least, and usually all – would have been absent as well.*" For its similarity to Hume, we confine our discussion in this paper to Lewis' definition.

From a structural engineering perspective, one may say the following; a given Beam A would deflect upon loading – implying that the application of loading triggers (or is the cause of) the deflection of Beam A. Similarly, one may also then infer the counterfactual that if it was not for the application of loading, Beam A would not have deflected. While these two examples may be perceived as logical, they as equally fundamental as they showcase the importance of causality in our domain.

How can we establish causality? To establish causality, we need to leverage causal models. Such models aim to describe (mathematically or otherwise) the data generating mechanism responsible for producing the observations we see. Before we dive into such models, let us start our discussion by demonstrating the broader concepts of causal discovery and causal inference necessary to realize causal models.

From this view, *causal discovery* (also referred to as causal structure search) seeks to discover the underlying structure of causal relations between variables pertaining to a DGP by analyzing observational data. Let us revisit Beam A. Say that this beam is simply supported, made from one homogenous material, has a depth, $D$, a width, $W$, a span, $L$, and deflects by a magnitude of $X$ upon the application of loading, $P$. If we carry out experiments where we intervene on $D$, $W$, $L$, and $P$, to observe how the deflection changes in response to our interventions, then we will be able to collect a series of observations pertaining specifically to each of our interventions. These observations can help us pinpoint the DGP responsible for the deflection of Beam A – so that we can: 1) predict the deflection of this beam without having to carry out costly/lengthy experiments, or 2) identify a suitable combination of interventions to limit the deformation of Beam A to a predefined limit. Causal discovery hopes to arrive at the *structure* behind the DGP that would allow us to answer the above two items.

A *causal structure* can be represented mathematically by causal models. A causal model spells out the probabilistic (in)dependence of variables and the effects of interventions (actual or hypothetical changes on one/some/all variables) [17]. Such a model can be further presented by a set of equations (namely, structural equation models (SEMs)) and/or accompanying graphs (i.e., Directed Acyclic Graphs[11] (DAGs)). An SEM represents assumed causal relationships between the involved parameters in an equational format that may entail a parametric or non-parametric form, while a DAG visually represents the causal structure of the DGP.

One should keep in mind that a DAG contains edged arrows that flow in one direction (e.g., $A \rightarrow B$, meaning $A$ causes $B$, etc.) and does not allow for a circular flow of information. Figure 5 shows an example of an SEM and a corresponding DAG. In this figure, $A$ is the direct cause of $B$, $C$ and $D$. $B$ is the direct cause of $C$. $A \rightarrow B \rightarrow C$ is called a path. In addition, A also serves as an indirect cause of $C$, given its role on $B$. Further, a node, say $B$, is a descendent of $A$ (and so on). DAGs are interpreted causally at the interventional and counterfactual levels, while are associational levels are used to describe conditional independencies between variables. There is more to SEM and

---

[11] A graph is a is a mathematical object that consists of nodes and edges. A DAG is a graph with directed edges. An example is shown in Fig. 5.

DAGs than the allowable page limit to this work. For additional sources, the reader is encouraged to review the following [18,19].

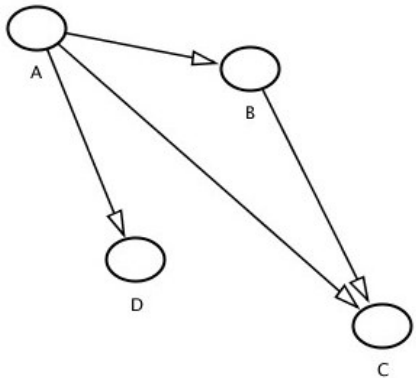

Fig. 5 Illustration of a DAG [corresponding SEMs are, $D := f_1(A)$, $B := f_2(A)$, $C := f_3(A, B)$. Note that error terms are assumed independent and are not explicitly shown in the figure. Also, $:=$ implies a causal statement]

As one can see from Fig. 5, there is a number of ways two parameters (or nodes) can be tied together. Other than direct causes and indirect causes, three main relationships are of interest to causal analysis: mediators, moderators, and colliders (see Fig. 6). To parallel cited works, this study will remain true to the terminology used in such works; hence, $X$ denotes a cause, $Y$ is an outcome (target), and Z is a third variable that could be a mediator, moderator, or a collider.

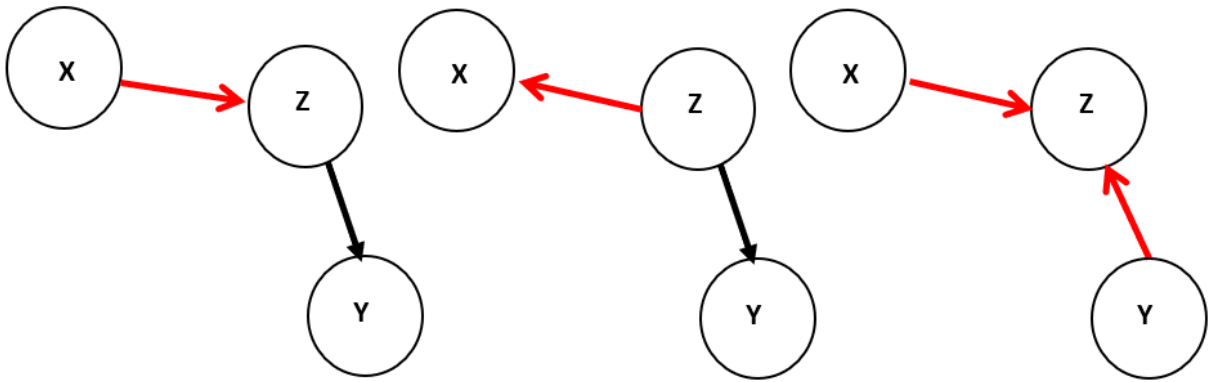

Fig. 6 Illustration of DAGs for mediators, confounders, and colliders (from left to right) [Note: the different arrowheads and colors are adopted for illustration purposes]

A mediator (or a chain) is a pattern in the form of $X \rightarrow Z \rightarrow Y$. This format implies that $X$ causally affects $Y$ through $Z$[12]. For example, fire ($X$) generates elevated temperatures ($Z$), which in turn degrade the sectional capacity of a load bearing member ($Y$). A companion to mediators is moderators, often defined as variables that change the *size* (or *direction*) of the relationship between variables. The following is an example of a moderator. The magnitude of deformation ($Y$) in a given beam can be amplified during fire conditions. In this instance, the influence of temperature rise (*Z*) can lead a beam to undergo larger deformations under the same level of loading (*X*). Such larger deformations may not be observed under ambient conditions from *X*. Thus, *Z* <u>amplifies</u> the deformation.

---

[12] For example, isolating $A \rightarrow B \rightarrow C$ from Fig. 5 into a new DAG can be considered a chain.

Reversing the arrow that extends from *Z* to *X* changes this pattern into a confounder (common cause or a fork). In this formation, *X* ← *Z* → *Y*, *Z* causally affects *X* and *Y*; effectively labeling *Z* as a confounder and inducing a *non-causal association* between *X* and *Y*. This formation can be tricky to investigate since a confounder can be a variable that we do not observe, do not often consider, do not know it exists, nor have data on. A hypothetical example would be the following: seismically active regions (*Z*) witness low structural failures (*X*) despite having a large number of structures (*Y*) than other regions, i.e., low structural failures (*X*) ← seismically active regions (*Z*) → high number of structures (*Y*). In this instance, the confounder *Z* makes *X* and *Y* statistically correlated despite the lack of a direct causal link between *X* and *Y*[13].

Finally, a collider turns up when the arrows of the moderator flip towards *Z*. Now, *Z* is no longer a cause but rather a common effect of *X* and *Y*. For example, harsh environmental conditions (*X*) and poor maintenance (*Y*) can lead to structural issues (*Z*) such as cracking and corrosion; thus, harsh environmental conditions (*X*) → structural issues (*Z*) ← poor maintenance (*Y*). If we are to condition on *Z*[14], then the association between *X* and *Y* in the mediator and confounder blocks the flow of association while doing the same on the collider induces non- causal association.

*Assumptions needed to establish causality*
There are several core assumptions tied to causal discovery and causal inference. These assumptions form the basis for establishing causality – especially from observational data. As such, these are described herein, and rich examples and corresponding mathematical details can be found elsewhere [8,17,20,21].

We start with the *causal Markov assumption*. This assumption state that a variable, *X*, is independent of each other variable (except *X*'s effects) conditional on its direct causes. For example, *X* in the left DAG in Fig. 6 is independent of *Y*, given (or conditional on) *Z*. A companion to the Markovian assumption is the *d-separation* criterion [22]. This criterion establishes whether *X* is independent of *Y*, given *Z* (i.e., $X \perp\!\!\!\perp Y \mid Z$), by associating the notion of independence with the separation of variables in a causal graph. For example, a path is d-separated if it contains 1) a chain or a fork such that the middle node is not in *Z*, or 2) a collider such that the middle node is not in *Z,* nor its descendants are in *Z*.

---

[13] An explanation would be that a seismically active region might home a metropolitan. Such a metropolitan would house a large population which requires the development of many structures. In addition, structures of seismically active regions are often required to comply with codal provisions that enforce additional detailing and requirement to mitigate damage.

[14] Building on the above discussion and referring to Fig. 6 while doing so may raise the question of how to distinguish a confounder from a collider (from the data or when domain knowledge is limited)? This task can be completed via the *back-door* and *front-door adjustments* [49]. In the confounding formation, *X* ← *Z* → *Y*, the back-door adjustment states that 1) no node in *Z* is a descendent of *X*, and 2) any path between *X* and *Y* that begins with an arrow into *X* (known as a back-door path) is blocked by *Z*, then controlling for *Z* blocks all non-causal paths between *X* and *Y*. In the event that the confounding factor is *unobservable* or *hypothetical*, then the front-door adjustment comes to the rescue. In this process, we add a new parameter that we may assume cannot be caused directly by the confounding factor. Now, we can use the aforementioned back-door adjustment to estimate the effect of the new parameter on the outcome. For an exhaustive description of the above two adjustments, please refer to the following [50,51].

*Causal faithfulness* implies that any population produced by a causal graph has the independence relations obtained by applying d-separation to it. Embracing this assumption eliminates all cases of unfaithfulness (i.e., independences that are not a consequence of the causal Markov condition or d-separation) from consideration. According to Heinze-Deml et al. [23], the causal Markov and faithfulness assumptions suggest that d-separation relationships in a causal DAG have a one-to-one correspondence with conditional independencies in the distribution of that particular graph. An example of unfaithfulness is one where *X* is a cause of *Y* and *Y* is a cause of *Z*, but *X* and *Z* are independent of each other [24].

Finally[15], we cover the c*ausal sufficiency* assumption. This assumption refers to the absence of hidden or latent parameters that we do not know nor are aware of, i.e., our data contains measurements on all of the common causes of the selected variables responsible for a DGP [20].

*Graphical methods*
Now that the fundamentals of causality are presented, it is time to move into the graphical methods that can be used to arrive at causal models and causal graphs[16].

Causal models have a causal structure and hence arriving at such models requires us to obtain the underlying structure of the problem on hand. While we may not know such a structure, we are likely to have data that we can exploit to arrive at a causal structure (which may turn out to be the ground truth or one that we are comfortable with labeling as a causal structure pending a series of assumptions arising from domain knowledge).

Causal search methods span different granularities – see Fig. 7. For example, a *directed graph (DG)* is a graph that allows loop feedback, and a *partially directed graph (PDG)* may contain both directed and undirected edges, while a DAG assumes acyclicity. In many instances, a causal DAG may not be identifiable, but rather the search method returns a set of equivalent DAGs. Such graphs are referred to as Markov equivalence (i.e., encode the same set of d-separation relationships). For example, *X*→*Z* →*Y, X* ←*Z* →*Y,* and *X*←*Z* ←*Y* contain the same conditional independence (i.e., *X* ⫫ *Y* | *Z*).

*Completed Partially Directed Acyclic Graphs (CPDAGs)* can be used to represent the above set of equivalent graphs. In a CPDAG, an edge is only directed if there is only one graph in the Markov equivalence class with an edge in that direction, and if there is uncertainty about the direction, this particular edge is left undirected. In the event that the search method identifies the presence of unobserved variables or confounders, such a method may return an *Acyclic Directed Mixed Graph (ADMG)*. This graph presents the unobserved variable with bidirected edges *X* ↔ *Y*; meaning there exists a possible confounder between *X* and *Y*. *Maximal Ancestral Graph (MAG)* have the capability to represent the same features of ADMGs as well as the presence of selection bias (preferential omission of data points from the samples [25]). *Partial Ancestral Graphs (PAGs)* are similar to CDPAGs as they also represent the set of equivalent MAGs (see Fig. 7 for an illustration of the aforenoted graphs).

---
[15] Other assumptions also exist such as Gaussianity of the noise distribution, one or several experimental settings, and linearity/nonlinearity, acyclicity – see [23].
[16] A thorough review on graphical model can be found in [24].

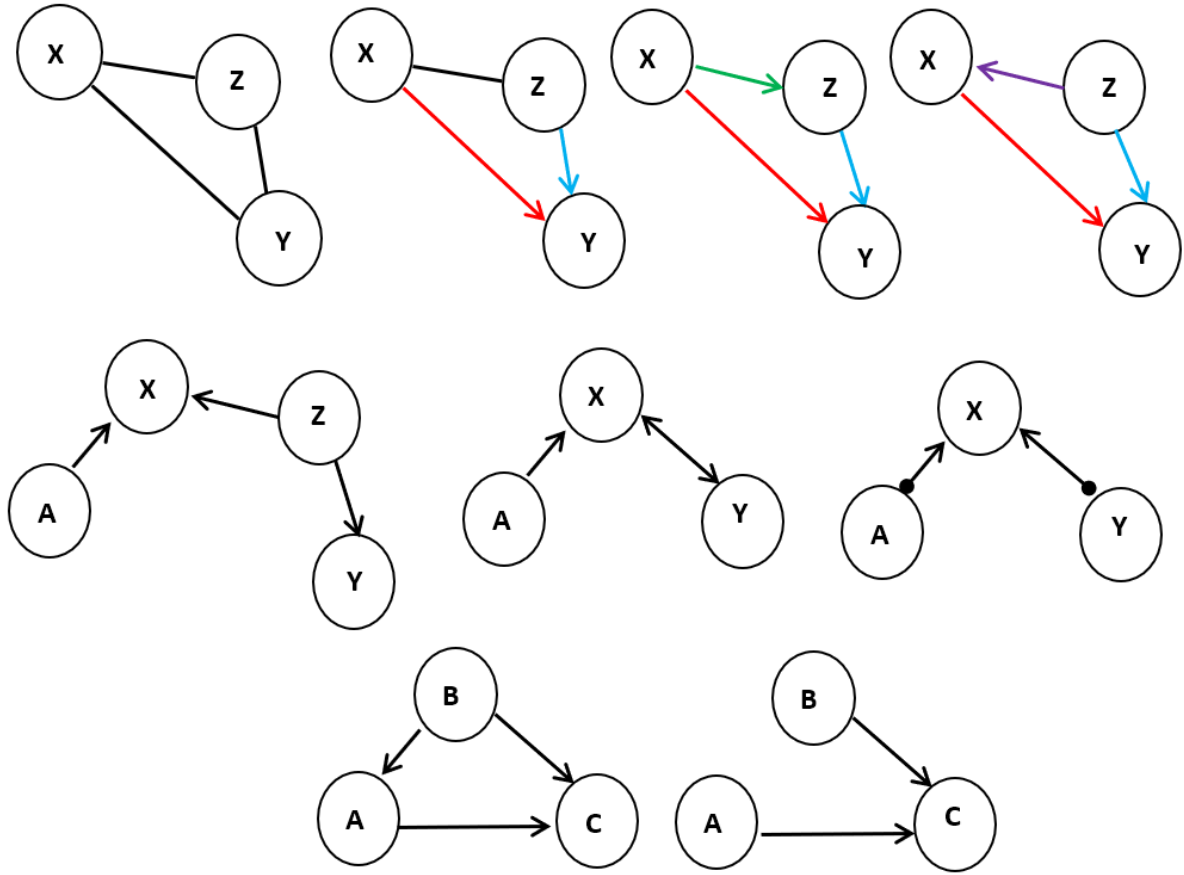

Fig. 7 Graphs [Note from top row and going right: graph (skeleton), CPDAG, DAG Markov equivalent 1, DAG Markov equivalent 2. Middle row left and going right: Assumed true DAG, MAG, and PAG. Bottom row: DAG, DAG with an intervention on *A* (i.e., do(*A*=*a*))]

Thus, the above graphs can home a number of relationships such as those outlined below. In principle, a causal search algorithm will attempt to tie the variables involved via such relationships, as discussed below.

*X→Y* where *X* is a cause of *Y*

*X – Y* where *X* is a cause of *Y* or vice versa

*X↔Y* where there is an unmeasured confounder

*X∘→Y* either *X* is a cause of *Y*, or there is a confounder

*X∘—∘Y* either *X* is a cause of *Y*, and/or vice versa, and/or there exists a confounder

*Causal search methods and causal machine learning packages*

There are two primarily search approaches to structure discovery[17]. These are *constraint-based* and *score-based*. This section covers key algorithms often used in each front, and other algorithms can be found elsewhere [23,26,27].

*Constraint-based* approaches arrive at causal structures by testing for constraints or conditional independencies as a means to construct a graph (i.e., DAG) that reflects and satisfies such constraints. In other words, these approaches seek to only find the graphs that correctly represent the independence relationships through hypothesis testing [17]. Additional rules are then applied to determine the relationship between each variable. As noted in the previous section, it is quite possible to find multiple graphs that fulfill a given set of conditional independencies. Therefore, constraint-based approaches are likely to output a graph representing Markov equivalence (CDPAGs or PAGs). Some of the commonly used constraint-based algorithms include Peter-Clark (PC) algorithm [28] and its variants (PC-stable, MPC, MRPC), Fast Causal Inference (FCI), and Inductive Causation (IC) [29].

On the other hand, *score-based* approaches identify graphs by assigning a relevance score (say, Bayesian Information Criterion) to candidate graphs. Since these approaches are expected to score every candidate graph, then they turn computationally expensive, and hence such approaches are often accompanied by greedy heuristics. Such approaches may include Greedy Equivalence Search (GES), Fast Greedy Equivalence Search (FGES), Greedy interventional equivalence search (GIES), etc.

While constraint-based approaches have been noted to be more efficient as they output one graph with clear semantics, they lack an indication of relative confidence in the output graph. On the other hand, score-based counterparts provide a series of equivalent graphs with an added confidence metric(s). Hence, researchers moved to explore hybrid approaches to maximize the output of causal searches [30]. Such a hybrid approach may include *structural agnostic modeling (SAM)* and *causal additive models (CAM)* [31].

In a similar manner to traditional ML algorithms, the outcome of a causal analysis can be evaluated via metrics (see Table 1). A number of metrics are available and include those traditionally used in classification problems [32,33] – especially if the ground truth mechanism (or one deduced from domain knowledge) is available. When the ground truth is not available, the performance of the causal discovery algorithm is evaluated on how well the causal structure predicts the phenomenon on hand.

---

[17] Vowels et al. [26] and Heinze-Deml [23] identify additional approaches such as causal association rules, causal forest, and causal networks. These were not presented here for simplicity. In addition, approaches pertaining to time series data, mixed data, or Bayesian methods were also left out.

Table 1 Metrics used for causal discovery analysis [17,23]

| Metric | Description |
| --- | --- |
| Missing or extra edges | Number of edges that are missing or present in the original model but not in the generated one. |
| Incorrect adjacencies | Number of undirected edges that are present in the generated model but not in the original one. |
| Incorrect and correct directed edges | Number directed edges that are missing present in the generated model that were correctly directed. |
| Structural hamming distance | Sum of missing edges, extra edges, and incorrectly directed edges. |
| Adjacency precision | Correctly predicted adjacencies/predicted adjacencies. |
| Adjacency recall | Correctly predicted adjacencies/true adjacencies. |
| Arrowhead precision | Correctly predicted arrowheads/predicted arrowheads. |
| Arrowhead recall | Correctly predicted arrowheads/true arrowheads. |
| Area-above-curve | AOC = 1−AUC, where AUC (area-under-curve) measures the area below the graph of false positive rate and true positive rate. |

*Causal inference and causal machine learning packages*
*Causal inference* seeks to study the possible effects of altering a given causal system. Simply put, causal inference infers how interventions, treatments, manipulations systematically alter the observations. To realize such inference, a *causal structure* that describes the phenomenon we are interested in is expected to be available. Going back to our example of a plate girder. Say that we were able to discover one possible causal structure for the DGP responsible for the corrosion of this plate girder. Then, using this discovered causal structure, we can answer questions about how altering the properties of this plate girder would influence its corrosion. In most interesting cases, we hope to infer the influence of a variable that we have not explored in an experiment.

We can also causally infer with regard to a certain phenomenon for which we do not know its causal structure. In this instance, the causal inference procedure would require the availability of data. Such data can be examined via a series of causal assumptions to realize their corresponding causal effects. These assumptions and rules are discussed herein[18].

In the Potential Outcome (PO) approach, as pioneered by Rubin [34], causal effects can be estimated by comparing the *potential outcomes* of a given intervention/treatment/decision. In reality, only one outcome can be observed since we cannot intervene on a unit and then un-intervene on the same unit at the same time. As such, causal inference is effectively a missing data problem, for which we can only observe one/some but not all of the outcomes. The potential outcomes are noted with $Y$. $Y_0$ notes the potential outcome for the untreated/benchmarked unit(s), and $Y_1$ notes the potential outcome for the treated/intervened upon unit(s). It is common to denote the intervention or treatment with a $T$.

---

[18] For simplicity and to provide our engineers with a wide perspective on causality, I will be adopting the Potential Outcome (PO) approach, instead of the Pearlian approach favored in causal discovery, here. Please note that both approaches present equivalent concepts [48]. A key concept in PO is the notion of randomization – often satisfied when setting upon experiments in structural engineering (i.e., we randomly assign intervention to load bearing specimens in the accompanying examples above).

Say, we are interested in inferring the benefits on flexural capacity obtained by casting RC beams using ultra-high performance concrete (UHPC) instead of traditional concrete, i.e., normal strength concrete (NSC). One way to carry out this investigation is by casting two identical groups of beams; Group A made from NSC ($Y_0$) and Group B made from UHPC ($Y_1$). Each group will home three identical beams (with the exception of concrete type) such that Group A [$A_{0,1}, A_{0,2}, A_{0,3}$], and Group B [$B_{0,1}, B_{0,2}, B_{0,3}$]. Then, we test all beams under bending and report our findings. To realize an answer (or a series of answers) to our question, we will need to causally infer such answer(s).

The simplest approach is to average the results of the flexural capacities of Group A and Group B and compare these averages (or expectations ($\mathbb{E}$)). Mathematically, this results in the following:

The difference in average (expectation) between Group A and Group B = $\mathbb{E}[Y_1 - Y_0]$ (2)

Equation 2 mirrors that which we could have arrived at using the Pearlian approach from intervention as noted by the do-operator (see Eq. 3)[19].

$$\mathbb{E}[Y|do(T=1)] - \mathbb{E}[Y|do(T=0)] \qquad (3)$$

Both equations 2 and 3 estimate the Average Treatment Effect (ATE). Additional causal effects also exist, such as the conditional average treatment effect (CATE)[20] such as the Average Treatment Effect on Treated (ATT) – what is the expected causal effect of the treatment for individuals in the treatment group?, as well as the Individual Treatment Effect of unit (ITE) – What is the expected causal effect of the treatment on the outcome of a specific unit? etc.

To continue our example in order to estimate ATT, we would want to go back to Group B only. Then, ATT would be the $\mathbb{E}[Y_1 - Y_0|T=1]$. It is obvious that ITE may not be easily nor possibly estimated from this example since we can only test each exact beam under one consideration. On a more positive note, the difference in expectation between each identical beam can give a close estimate of ITE.

The PO approach extends further than the limited space of this work to cover a wide range of applications and conditions (i.e., propensity scoring, confoundness, missing data, matching, etc.), which can be reviewed herein [35]. There are a number of ML packages that can be used in causal inference investigations.

**Case Studies**

This section describes three case studies to be highlighted in this work. The first two revolve around causal discovery (via constraint-based and score-based algorithms)[21], and the third covers causal inference.

---

[19]Note that $p(Y = y| do(X = x)) = p_m(Y = y | X = x) = \sum_z (Y = y | X = x)p(Z = z)$, where $p_m$ is the manipulated distribution arising from our intervention. A review on do-calculus can be found in [52], and a sample solved example is shown at a later section.

[20] Simply, an ATE conditioned on a subset of the population.

[21] The used algorithms are the PC and FGES in their default settings (given the illustrative nature of this study). For more information on these algorithms, please refer to their original sources [23,53].

*Causal discovery*

Case 1: Obtaining the causal structure for a known DGP

The first case study hopes to verify our domain knowledge of a known and simple DGP. The DGP of interest pertains to the moment capacity, *M*, of W-shaped steel beams, which was discussed in Eq. 1. As mentioned earlier, the moment capacity of such beams can be estimated by multiplying the plastic modulus, *Z*, with the yield strength of steel, $f_y$; effectively labeling *M* as a collider, $Z \rightarrow M \leftarrow f_y$. In other words, *M* is the effect of *Z* and $f_y$ – a similar DAG for a collider is shown in Fig. 6.

Let us explore two algorithms (namely, the PC algorithm, a *constraint-based,* and the FGES algorithm, a *score-based*). To execute this analysis, a database comprising all W-shaped structural steel sections adopted from the American Institute of Steel Construction (AISC) manual [36] was developed. In this database, the plastic modulus of each steel section was multiplied by two steel Grades (A992 and A36). Therefore, this data contains three variables, two independent (*Z* and $f_y$) and one dependent (*M*). The compiled database was used in this analysis, and the results of this analysis are shown in Fig. 8. As one can see, the generated DAG resembles that of our domain knowledge and is identical to both algorithms. This is both expected and comforting. As mentioned, only score-based algorithms are fitted with the capability to fit score-of-fitness to the developed DAG. This score was -19236.4 based on the default Bayesian Information Criterion (BIC)[22].

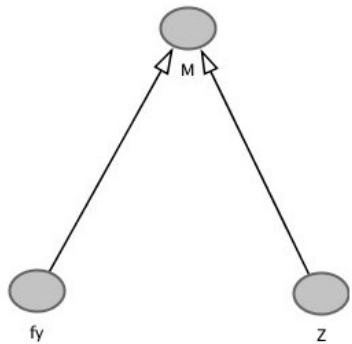

Fig. 8 Result from case study no. 1 successfully resembling a collider

Case 2: Obtaining the causal structure for an unknown DGP

In this case study, we are interested in uncovering the unknown DGP of how RC columns spall under fire conditions. In this pursuit, and to demonstrate some critical differences between causal discovery analysis and a traditional ML analysis, a particular database of fire-exposed RC columns, which has been explored via traditional and explainable ML algorithms in an earlier work of the author [37], was incorporated herein. Thus, the presented case study is designed to identify the DGP for spalling in RC columns and contrast the causal analysis to a recently published ML analysis.

---

[22] BIC score is calculated as chi-square statistics - degrees of freedom × log(sample size). Low values of BIC imply the goodness of the model.

The compiled database includes the following variables: column width and depth, $b$, steel reinforcement ratio, $r$, concrete compressive strength, $f$, concrete cover to reinforcement, $C$, magnitude of applied loading, $P$, and Spalling, $SP$ (Yes, or No). A visual illustration of the aforenoted variables via histograms, along with additional descriptive statistical details, are shown in Table 2 and Fig. 9. It is worth noting that the former analysis conducted via an explainable ensemble has identified the following five variables as the most important in terms of their influence on fire-induced spalling of RC columns: $C$ (100%), $f$ (95%), $b$ (88%), $r$ (66%), and $P$ (44%).

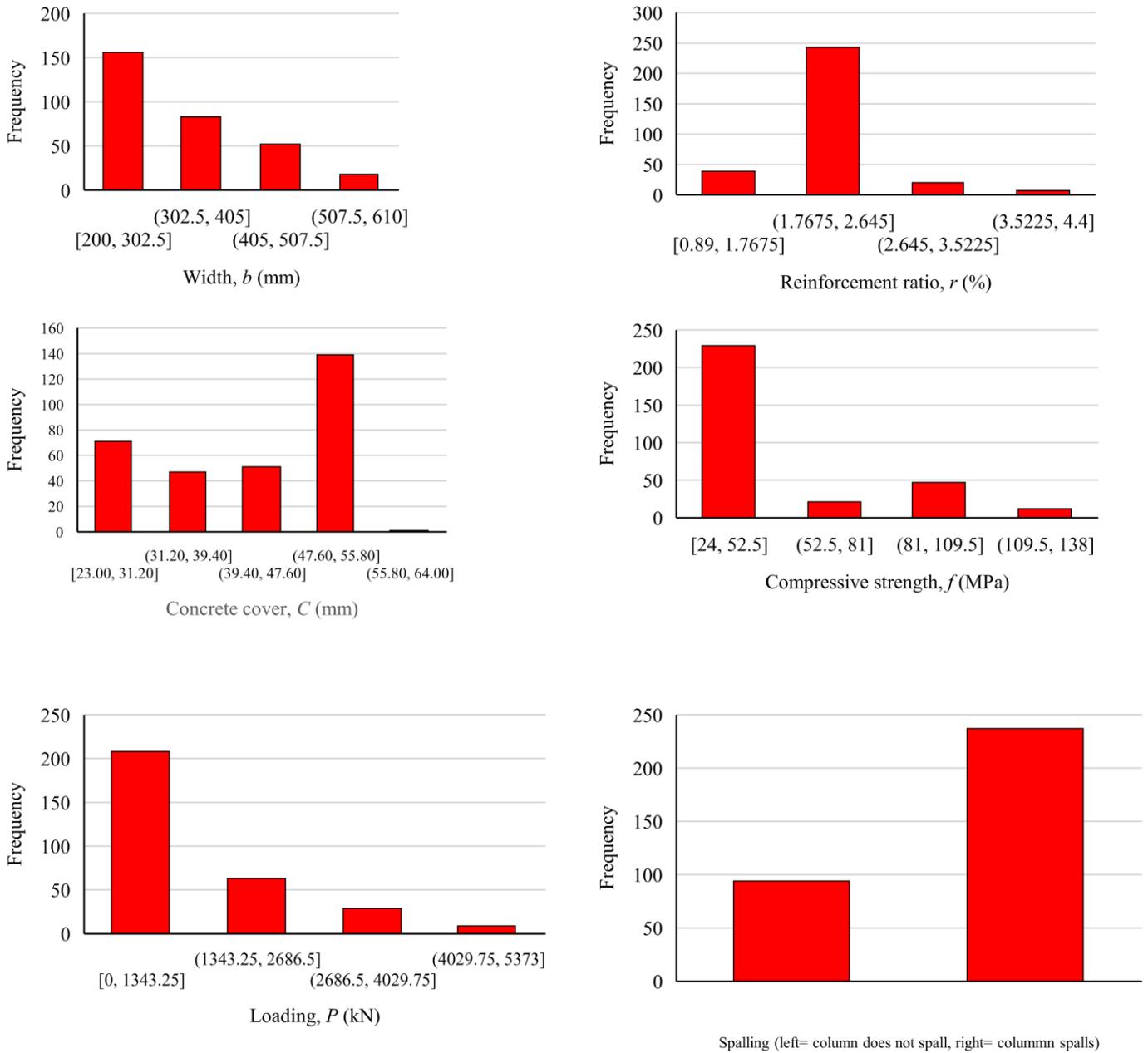

Fig. 9 Frequency of identified features of selected RC columns in the compiled database

Table 2 Statistics on collected database

|  | b (mm) | r (%) | f (MPa) | K | C (mm) | P (kN) |
|---|---|---|---|---|---|---|
| Minimum | 152.0 | 0.3 | 15.0 | 0.5 | 13.0 | 0.0 |
| Maximum | 514.0 | 11.7 | 126.5 | 1.0 | 64.0 | 5373.0 |
| Average | 325.0 | 2.3 | 42.0 | 0.5 | 33.4 | 1335.2 |
| Standard deviation | 72.6 | 1.5 | 24.2 | 0.1 | 7.7 | 999.1 |
| Skewness | 0.7 | 2.5 | 1.9 | 3.6 | -0.4 | 1.6 |

Similar to the first case study, the constraint-based PC algorithm and the score-based FGES algorithm were applied to identify the causal structure of the DGP. The outcome of this analysis is shown in Fig. 10. This figure shows two possible structures per algorithm. The first (left) is obtained by using all the variables (i.e., each algorithm will have to determine the relationships between variables as well as which variable is the outcome of interest, and the second (right) is obtained by identifying spalling as the outcome and acknowledging that none of the independent variables can cause each other.

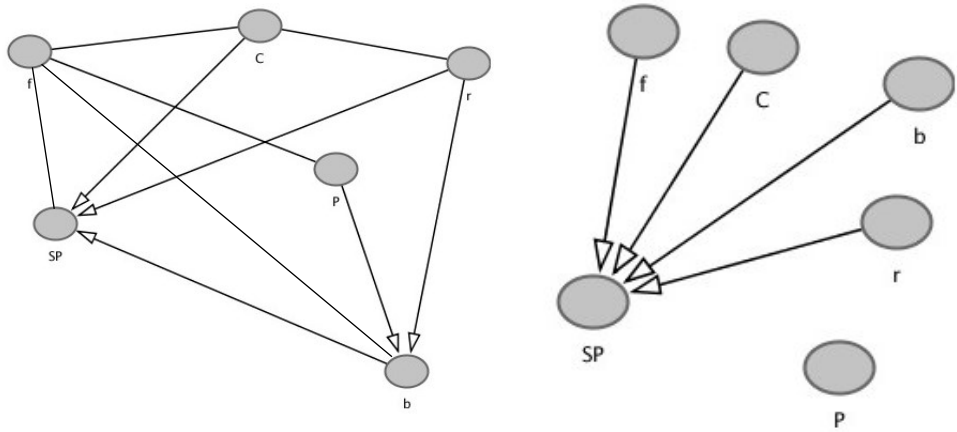

(a) The outcome from the PC algorithm

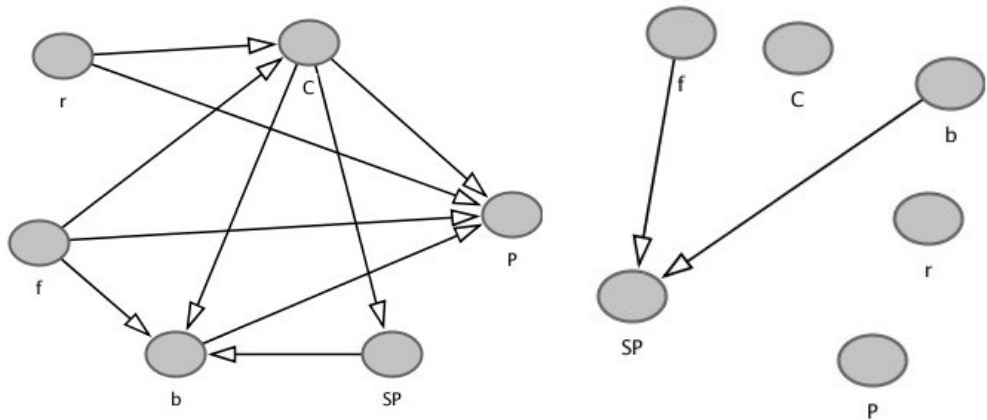

(b) The outcome from the FGES algorithm (BIC = 0.0)
Fig. 10 Comparison between DGP [Note: left without domain knowledge, right: with domain knowledge]

Figure 10(right) shows that the PC algorithm was not able to create a DAG, while the FGES algorithm managed to do so successfully. A look into these figures also shows that the PC identified *C*, *r*, *f*, and *b* with possible relations (causal from *C*, *f*, and *b*, as well as non-causal from *f*). This algorithm does not relate *P* to spalling. On the other hand, the FGES only notes a direct causal relation between *C* to spalling. It is clear that both of these figures do not provide us with sufficient information to uncover a likely DGP. Moving to Fig. 10(left), this figure shows that PC and FGES manage to create a DAG each. These two DAGs do not agree on the DGP. Of all the DAGs, the DAG created by PC(left) is the closest we can get to our domain knowledge as obtained from notable studies on spalling that <u>qualitatively</u> paint a picture of the possible DGP behind spalling [38–41]. This causal analysis may shed some grey over the suitability of the former ML analysis (as that particular analysis cannot be assumed to uncover the causality of spalling but rather a prediction tool from the first rung of the causal ladder). Hence, a more in-depth analysis of such DGP is better reserved for a dedicated study that is under exploration at the moment[23].

*Causal inference*

<u>Case 3: Inferring the benefits of alternative systems</u>

The following is a case study about inferring the causal effects of whether the adoption of a new strengthening system (i.e., made from fiber-reinforced polymer composites, $T = 1$) is more beneficial than an existing system (e.g., one crack filler, $T = 0$) on the capacity of RC beams ($C$). Data were collected from two branches of the same firm (Branch A and Branch B) to be analyzed to decide if this firm can adopt the new system across all of its branches. To do so, we need to estimate the causal effect of the treatment ($T$) on the sectional capacity as listed in Table 3.

Table 3 Data used in the causal inference example (total beams = 716)

| Element | New system ($T = 1$) | Old system ($T = 0$) |
|---|---|---|
| Branch A | 81 out of 87 improved (93%) | 243 out of 280 improved (87%) |
| Branch B | 190 out of 262 improved (73%) | 60 out of 87 improved (69%) |
| Totals | 271 out of 349 improved (78%) | 303 out of 367 improved (83%) |

A look into Table 3 shows that the new system outperforms the old system in both branches individually (93% vs. 87%, and 73% vs. 69%, respectively). A second glance at this table indicates that the same system underperforms when compared in terms of the totals (78% vs. 83% for Branch A and Branch B, respectively). In all cases, it is unclear how the collected data can be interpreted and analyzed. This intriguing data, as homed in Table 3, is an example of Simpson's Paradox [42,43][24] where a proper causal inference investigation is warranted to estimate the true causal effects.

The chief engineer ($E$) is a proponent of the new system. Thus, being the chief engineer has an effect on adopting the new system ($T = 1$), which in turn has an effect on sectional capacity ($C$). This can be translated into the DAG to the left in Fig. 11.

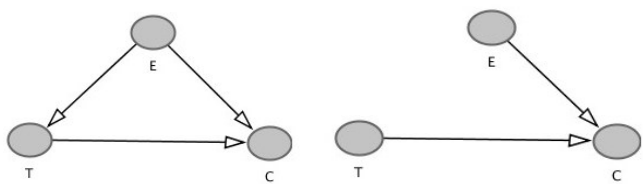

Fig. 11 DAG for the case study on hand

---

[23] Another reason to not delve into the DGP in this study is that preliminary analysis conducted using other algorithms (i.e., RFCI and FAS) identified the presence of confounders. Tackling such confounders is hectic and is best left for a future work.

[24] Please note that this is a hypothetical example that stems from those showcased in the works of Pearl [13] and Fabian [14].

Fundamentally, this firm is interested in the probability of improvement if both branches are forced to adopt or not adopt the new system. In other words, the difference between these two probabilities is indeed the average causal effect in the population of RC beams listed in Table 3. This estimate resembles that which can be arrived at via Eqs. 2 and 3.

We will need to then calculate the $p(C = 1| do(T = 1))$ and $p(C = 1 | do(T = 0))$ which can be arrived at from the DAG to the right of Fig. 11. For simplicity, these calculations are shown herein:

$p(C = 1| do(T = 1)) = p(C = 1 | T = 1, E = 0) p(E = 0) + p(C = 1 | T = 1, E = 1) p(E = 1)$

$= 81/87 \times (87+280)/716 + 190/262 \times (262+87)/716 = 0.83$

And,

$p(C = 1| do(T = 0)) = p(C = 1 | T = 0, E = 0) p(E = 0) + p(C = 1 | T = 0, E = 1) p(E = 1)$

$= 243/280 \times (87+280)/716 + 60/87 \times (262+87)/716 = 0.78$

Then, using Eq. 3 yields:

$\mathbb{E}[Y|do(T=1)] - \mathbb{E}[Y|do(T=0)] = 0.83 - 0.78 = 0.05$.

Please note that this estimate (+ 5%) infers that, on average, 5% more RC beams would benefit if retrofitted using the new system. Surprisingly, this is opposite to the conclusion one might have drawn from the aggregated data in Table 3.

**A Note on Causality and Traditional Machine Learning**
This section presents additional insights and observations that arose during this work, which may spark future discussion on causality in our domain.

The bulk of ML works published in our area heavily rely on the notion of predictive models. A linear approach is adopted in which data is first collected and then fed into an ML algorithm(s) for training and testing. The model is deemed *fit* once it satisfies a set of criteria and performance indicators. Revisiting Fig. 4 raises a series of questions. First, a traditional ML model aims to predict a value for an outcome. This prediction is likely to arise from a non- causal procedure since the ML model, in principle, was not built using causal principles.

Second, the ML model treats all inputs from a data-driven perspective. As such, this model may alter such inputs (engineer features) in order to balance the required precision and/or simplicity of the model (among other items such as energy trade-offs, etc.). Most importantly, a typical ML model maximizes the variance between predictions and ground truth and assumes that the data points are *independent and identically distributed* (iid)[25]. The aforenoted is a common assumption in ML and is critical to be satisfied. Moreover, ML models rarely check for confoundness and selection bias.

---

[25] Each data point was generated without reference to others, and the underlying distributions in the DGP are the same for all of the points.

On the other hand, a causal model hopes to estimate a causal effect, and hence its fitness arises from the accuracy of estimates as obtained from the causal model. Such a model emphasizes the role of the input variables and seeks evidence (domain knowledge and judgment[26]) to identify such a role in establishing causality, minimizing/eliminating confoundness, and selection bias. An important note to remember is that a causal model is interpretable by nature as it conveys the DGP; however, a ML may not be[27].

While this study is a proponent of causality, it is worth noting that we may not need to truly establish causality and build causal models for all of our problems. In some problems, data-driven model continues to be of aid. In others, we do need to have interpretable traditional ML models. In other problems, we may not be in need of knowing the DGP as our domain knowledge can overcome such need! It is also equally important to remember that establishing causality is challenging, and we may lack the data or tools to overcome such challenges. Perhaps future advancements in the coming years will help us find new ways to realize such a goal.

**Conclusions**

This work presents big ideas and concepts behind causality, causal discovery, and causal inference from a civil and structural engineering lens. The primary motivation of this study is to showcase the merit in adopting causal principles to identify data generating processes we happen to be interested in to further our understanding of such processes. In this pursuit, two of the main approaches to causality (the Pearlian approach and the Potential Outcome approach) are discussed and then examined via two case studies. The following list of inferences can also be drawn from the findings of this study:

- Causality is an exciting and uncharted area of research, and our domain is expected to significantly thrive from future success in this area.
- Identifying the actual data generating process is a natural pursuit for discovering new knowledge.
- There is a dire need to integrate causal principles into traditional ML methods as a means to maximize the output of such methods.
- A variety of causal discovery and causal inference algorithms and packages exist. Yet, none seem to be created by civil engineers.

**Data Availability**

Some or all data, models, or code that support the findings of this study are available from the corresponding author upon reasonable request.

**Acknowledgment**

I would like to thank Eng. Arash Tapeh for his help in preparing Figure 1, as well as Johannes Textor for creating DAGitty.

---

[26] For transparency, some ML models also apply domain knowledge to justifying the rationale behind selecting or neglecting inputs.

[27] There could be a debate with regard to explainable AI (XAI). However, XAI primarily explains the reasoning behind a prediction from a data perspective, and not with regard to DGP. Thus, at a fundamental level, an XAI model may still remain a blackbox since our data may not be complete, may include confounders, etc.

**Conflict of Interest**

The author declares no conflict of interest.

**References**

[1] M. Bunge, Causality and modern science, 2017. https://doi.org/10.4324/9781315081656.

[2] B. Schölkopf, Causality for Machine Learning, (2019). https://doi.org/10.1145/3501714.3501755.

[3] Dimensions, Dimensions.ai, (2021). https://www.dimensions.ai/.

[4] M. Thelwall, Dimensions: A competitor to Scopus and the Web of Science?, J. Informetr. (2018). https://doi.org/10.1016/j.joi.2018.03.006.

[5] M.A. Beyzatlar, M. Karacal, H. Yetkiner, Granger-causality between transportation and GDP: A panel data approach, Transp. Res. Part A Policy Pract. (2014). https://doi.org/10.1016/j.tra.2014.03.001.

[6] T. Tong, T.E. Yu, Transportation and economic growth in China: A heterogeneous panel cointegration and causality analysis, J. Transp. Geogr. (2018). https://doi.org/10.1016/j.jtrangeo.2018.10.016.

[7] A. Gibb, H. Lingard, M. Behm, T. Cooke, Construction accident causality: Learning from different countries and differing consequences, Constr. Manag. Econ. (2014). https://doi.org/10.1080/01446193.2014.907498.

[8] N. Huntington-Klein, The Effect : An Introduction to Research Design and Causality, Chapman and Hall/CRC, Boca Raton, 2021. https://doi.org/10.1201/9781003226055.

[9] D.F. Chambliss, R.K. Schutt, Causation and experimental design., in: Mak. Sense Soc. World Methods Investig., 2013. https://us.sagepub.com/sites/default/files/upm-assets/103318_book_item_103318.pdf.

[10] K. Wardhana, F.C. Hadipriono, Analysis of Recent Bridge Failures in the United States, J. Perform. Constr. Facil. 17 (2003) 144–150. https://doi.org/10.1061/(ASCE)0887-3828(2003)17:3(144).

[11] K. Wardhana, F.C. Hadipriono, Study of Recent Building Failures in the United States, J. Perform. Constr. Facil. 17 (2003) 151–158. https://doi.org/10.1061/(ASCE)0887-3828(2003)17:3(151).

[12] Surveys | NCSES | NSF, (2022). https://www.nsf.gov/statistics/surveys.cfm (accessed March 26, 2022).

[13] M.D. Pearl J, The Book of Why: The New Science of Cause and Effect-Basic Books, Basic Books, 2018.

[14] F. Dablander, An Introduction to Causal Inference, (n.d.). https://doi.org/10.31234/OSF.IO/B3FKW.

[15] H.F. Klemme, Hume, David: A Treatise of Human Nature, in: Kindlers Lit. Lex., 2020. https://doi.org/10.1007/978-3-476-05728-0_9644-1.

[16] D. Lewis, Causation, J. Philos. 70 (1973) 556. https://doi.org/10.2307/2025310.

[17] A.R. Nogueira, A. Pugnana, S. Ruggieri, D. Pedreschi, J. Gama, Methods and tools for causal discovery and causal inference, Wiley Interdiscip. Rev. Data Min. Knowl. Discov. 12 (2022) e1449. https://doi.org/10.1002/WIDM.1449.

[18] K.A. Bollen, J. Pearl, Eight Myths About Causality and Structural Equation Models, in: 2013. https://doi.org/10.1007/978-94-007-6094-3_15.

[19] A. Forney, S. Mueller, Causal Inference in AI Education: A Primer, J. Causal Inference. (n.d.) 2021.

[20] R. Scheines, An Introduction to Causal Inference *, (n.d.).

[21] J. Pearl, Causal Diagrams and the Identification of Causal Effects, in: Causality, 2013. https://doi.org/10.1017/cbo9780511803161.005.

[22] J. Pearl, Causal inference in statistics: An overview, Stat. Surv. (2009). https://doi.org/10.1214/09-SS057.

26
[23] C. Heinze-Deml, M.H. Maathuis, N. Meinshausen, Causal Structure Learning, Annu. Rev. Stat. Its Appl. (2018). https://doi.org/10.1146/annurev-statistics-031017-100630.

[24] G.I. Allen, Handbook of Graphical Models, J. Am. Stat. Assoc. (2020). https://doi.org/10.1080/01621459.2020.1801279.

[25] E. Bareinboim, J. Tian, J. Pearl, Recovering from selection bias in causal and statistical inference, in: Proc. Natl. Conf. Artif. Intell., 2014. https://doi.org/10.1145/3501714.3501740.

[26] M.J. Vowels, N.C. Camgoz, R. Bowden, D'ya like DAGs? A Survey on Structure Learning and Causal Discovery, (2021). https://doi.org/10.48550/arxiv.2103.02582.

[27] A.R. Nogueira, J. Gama, C.A. Ferreira, Causal discovery in machine learning: Theories and applications, J. Dyn. Games. (2021). https://doi.org/10.3934/jdg.2021008.

[28] P. Spirtes, C. Glymour, R. Scheines, Causation, prediction, and search (Springer lecture notes in statistics), Lect. Notes Stat. (2000).

[29] K. Yu, J. Li, L. Liu, A Review on Algorithms for Constraint-based Causal Discovery, (2016). https://doi.org/10.48550/arxiv.1611.03977.

[30] S. Triantafillou, I. Tsamardinos, Score based vs constraint based causal learning in the presence of confounders, in: CEUR Workshop Proc., 2016.

[31] P. Spirtes, K. Zhang, Causal discovery and inference: concepts and recent methodological advances, Appl. Informatics. (2016). https://doi.org/10.1186/s40535-016-0018-x.

[32] H. M, S. M.N, A Review on Evaluation Metrics for Data Classification Evaluations, Int. J. Data Min. Knowl. Manag. Process. (2015). https://doi.org/10.5121/ijdkp.2015.5201.

[33] M.Z. Naser, · Amir, H. Alavi, Error Metrics and Performance Fitness Indicators for Artificial Intelligence and Machine Learning in Engineering and Sciences, Archit. Struct. Constr. 2021. 1 (2021) 1–19. https://doi.org/10.1007/S44150-021-00015-8.

[34] D.B. Rubin, Causal Inference Using Potential Outcomes, J. Am. Stat. Assoc. (2005). https://doi.org/10.1198/016214504000001880.

[35] G.W. Imbens, D.B. Rubin, Causal inference: For statistics, social, and biomedical sciences an introduction, 2015. https://doi.org/10.1017/CBO9781139025751.

[36] AISC, AISC Shapes Database v15.0H , Am. Inst. Steel Constr. Database. (2022). https://www.aisc.org/search/?query=shapes database&pageSize=10&page=1 (accessed March 30, 2022).

[37] M.Z. Naser, V.K. Kodur, Explainable machine learning using real, synthetic and augmented fire tests to predict fire resistance and spalling of RC columns, Eng. Struct. 253 (2022) 113824. https://doi.org/10.1016/J.ENGSTRUCT.2021.113824.

[38] K.D.D. Hertz, Limits of spalling of fire-exposed concrete, Fire Saf. J. 38 (2003) 103–116. https://doi.org/10.1016/S0379-7112(02)00051-6.

[39] G.A. Khoury, Effect of fire on concrete and concrete structures, Prog. Struct. Eng. Mater. 2 (2000) 429–447. https://doi.org/10.1002/pse.51.

[40] V.K.R. Kodur, Spalling in High Strength Concrete Exposed to Fire: Concerns, Causes, Critical Parameters and Cures, in: Adv. Technol. Struct. Eng., American Society of Civil Engineers, Reston, VA, 2000: pp. 1–9. https://doi.org/10.1061/40492(2000)180.

[41] G. Sanjayan, L. Stocks, Spalling of high-strength silica fume concrete in fire, ACI Mater. J. (1993). https://www.concrete.org/publications/internationalconcreteabstractsportal/m/details/id/4015 (accessed April 8, 2019).

[42] C.H. Wagner, Simpson's paradox in real life, Am. Stat. (1982).

https://doi.org/10.1080/00031305.1982.10482778.

[43] C.R. Blyth, On simpson's paradox and the sure-thing principle, J. Am. Stat. Assoc. (1972). https://doi.org/10.1080/01621459.1972.10482387.

[44] A. Michotte, The perception of causality, 2017. https://doi.org/10.4324/9781315519050.

[45] W.C. Salmon, Causality and Explanation, 2003. https://doi.org/10.1093/0195108647.001.0001.

[46] P.W. Holland, Statistics and causal inference, J. Am. Stat. Assoc. (1986). https://doi.org/10.1080/01621459.1986.10478354.

[47] C. Glymour, K. Zhang, P. Spirtes, Review of causal discovery methods based on graphical models, Front. Genet. (2019). https://doi.org/10.3389/fgene.2019.00524.

[48] G.W. Imbens, Potential outcome and directed acyclic graph approaches to causality: Relevance for empirical practice in economics, J. Econ. Lit. (2020). https://doi.org/10.1257/JEL.20191597.

[49] M.D. Pearl J, The Book of Why_ The New Science of Cause and Effect-Basic Books, 2018.

[50] A.N. Glynn, K. Kashin, Front-Door Versus Back-Door Adjustment With Unmeasured Confounding: Bias Formulas for Front-Door and Hybrid Adjustments With Application to a Job Training Program, J. Am. Stat. Assoc. (2018). https://doi.org/10.1080/01621459.2017.1398657.

[51] J.D. Correa, E. Bareinboim, Causal effect identification by adjustment under confounding and selection biases, in: 31st AAAI Conf. Artif. Intell. AAAI 2017, 2017.

[52] J. Pearl, CAUSALITY, 2nd Edition, 2009, Cambridge University Press, 2009. http://bayes.cs.ucla.edu/BOOK-2K/ (accessed March 28, 2022).

[53] J. Ramsey, M. Glymour, R. Sanchez-Romero, C. Glymour, A million variables and more: the Fast Greedy Equivalence Search algorithm for learning high-dimensional graphical causal models, with an application to functional magnetic resonance images, Int. J. Data Sci. Anal. (2017). https://doi.org/10.1007/s41060-016-0032-z.